\documentclass{article}

\usepackage[final]{neurips_2025}
\usepackage{graphicx}
\usepackage{amsmath}
\usepackage{wrapfig}
\usepackage{array}
\usepackage{multirow}
\usepackage{colortbl}

\usepackage[utf8]{inputenc} 
\usepackage[T1]{fontenc}    
\usepackage{hyperref}       
\usepackage{url}            
\usepackage{booktabs}       
\usepackage{amsfonts}       
\usepackage{nicefrac}       
\usepackage{microtype}      
\usepackage{xcolor}         

\definecolor{lightgreen}{RGB}{232,245,233}

\title{How Intrinsic Motivation Shapes Learned Representations in Decision Transformers: A Cognitive Interpretability Analysis}

\author{
  Leonardo Guiducci\\
  DISPOC\\
  University of Siena, Siena (Italy)\\
  \texttt{leonardo.guiducci@unisi.it} \\
   \And
  Antonio Rizzo\\
 DISPOC\\
  University of Siena, Siena (Italy)\\
  \texttt{antonio.rizzo@unisi.it} \\
  \And
  Giovanna Maria Dimitri\\
  University of Milan, Milan (Italy)\\
  \texttt{giovanna.dimitri@unimi.it} \\
}

\begin{document}

\maketitle

\begin{abstract}
Elastic Decision Transformers (EDTs) with intrinsic motivation have demonstrated improved performance in offline reinforcement learning, yet the cognitive mechanisms underlying these improvements remain unexplored.
We introduce a systematic post-hoc explainability framework to analyze how intrinsic motivation shapes learned embeddings in EDTs through statistical analysis of embedding properties (covariance structure, vector magnitudes, and orthogonality). We reveal that different intrinsic motivation variants create fundamentally different representational structures: one variant operating on state embeddings promotes compact representations, while another operating on transformer outputs enhances representational orthogonality. Our analysis demonstrates strong environment-specific correlation patterns between embedding metrics and performance across locomotion tasks.
These findings show that intrinsic motivation operates as a representational prior that shapes embedding geometry in cognitively plausible ways, creating environment-specific organizational structures that facilitate better decision-making beyond simple exploration enhancement.


\end{abstract}

\section{Introduction and Background}
\label{sec:Introduction}

Reinforcement learning has evolved beyond reactive policy optimization to include models with greater generalization and adaptability, particularly in offline reinforcement learning where agents learn optimal policies solely from previously collected data without environment interaction during training \cite{levine2020offline,kidambi2020morel,kumar2020conservative}. Elastic Decision Transformers (EDTs) \cite{wu2023elastic} have emerged as a promising architecture that unifies sequence modeling with decision-making by leveraging Transformer \cite{transformer} architectures to capture long-range dependencies and enable flexible policy behaviors under uncertainty. EDTs enhance standard Decision Transformers \cite{chen2021decision} through dynamic history length adjustment, enabling effective trajectory stitching for improved offline RL performance.

Intrinsic motivation mechanisms, inspired by cognitive science theories of curiosity and novelty-seeking behavior \cite{oudeyer2007intrinsic,pathak2017curiosity}, have been incorporated into RL to encourage exploration when extrinsic rewards are sparse or poorly aligned with long-term success. Recent work has shown that integrating intrinsic rewards into EDTs yields improved performance across offline RL benchmarks \cite{esann2025}. However, while the performance benefits are empirically established, the representational mechanisms underlying these improvements remain largely unexplored. Understanding how intrinsic motivation shapes the internal embedding spaces of EDTs is crucial for interpretable reinforcement learning, as these models learn implicit state representations in high-dimensional spaces that lack the interpretability of traditional hand-crafted features \cite{bengio2013representation}.

This paper addresses the gap between empirical performance gains and mechanistic understanding by investigating how intrinsic motivation shapes learned representations in EDTs. We introduce a systematic post-hoc explainability framework using statistical analysis of embedding properties (covariance trace, L2 norm, cosine similarity) to examine representational geometry. Our analysis reveals that intrinsic motivation operates beyond simple exploration bonuses, acting as a representational prior that creates environment-specific organizational structures. We analyze the two EDT variants presented in \cite{esann2025}: \textit{EDT-SIL}, where intrinsic loss operates on embedded states promoting compactness, and \textit{EDT-TIL}, where it operates on transformer outputs enhancing orthogonality. 
Our contributions include:
\begin{enumerate}
    \item Post-hoc explainability framework for analyzing embedding geometry changes.
    \item Mechanistic analysis revealing distinct representational structures of EDT variants.
    \item Demonstration of quantitative correlations between embedding properties and task performance across environments.
\end{enumerate}


\section{Methods}
\label{sec:Methods}

We analyze Elastic Decision Transformers enhanced with intrinsic motivation mechanisms \cite{esann2025}, building upon the EDT architecture \cite{wu2023elastic} that processes trajectories as sequences of (state, action, reward) tuples. Our analysis focuses on two intrinsically-motivated EDT variants that incorporate Random Network Distillation (RND) \cite{rnd} modules as auxiliary loss functions.

\subsection{Intrinsic Motivation Variants}
\label{subsec: intrinsi auxiliary loss}

We examine two EDT variants that differ in where the intrinsic signal operates: \textbf{EDT-SIL (State Input Loss)} computes intrinsic loss directly from embedded state representations, allowing the intrinsic signal to influence the state embedding layer and potentially encourage more structured representations. \textbf{EDT-TIL (Transformer Input Loss)} operates on transformer output representations, enabling the intrinsic signal to shape both embedding and transformer layers for more coherent sequential representations.

The intrinsic loss is computed as $L_{\text{int}} = |f_{\text{pred}}(x; \theta_{\text{pred}}) - f_{\text{target}}(x; \theta_{\text{target}})|_2^2$ where $x$ represents either embedded states (SIL) or transformer outputs (TIL). The total loss combines the standard EDT objective with this intrinsic component: $L_{\text{overall}} = L_{\text{EDT}} + L_{\text{int}}$. This formulation enables intrinsic motivation to enhance representation learning without disrupting the primary task objective.

\subsection{Post-Hoc Explainability Framework}
\label{subsec: post-hoc explainability analysis}

Our primary contribution is a framework for analyzing how intrinsic motivation shapes learned representations by examining geometric and statistical properties of embedding spaces. We focused on three key metrics that capture different aspects of representational structure: \textit{covariance trace} measuring total variance distribution across embedding dimensions, \textit{L2 norm} quantifying representational compactness, and \textit{cosine similarity} assessing representational orthogonality. 

To establish quantitative relationships between representational properties and task performance, we computed Pearson correlations between embedding metrics and normalized performance scores across multiple seeds, identifying the most predictive metric for each environment-model combination. This analysis reveals how different intrinsic motivation mechanisms create distinct representational patterns and provides mechanistic insights into why intrinsic motivation improves policy learning.



\section{Experiments and Results}
\label{sec:ExperimentsAndResults}

We evaluate intrinsic motivation mechanisms in Elastic Decision Transformers, focusing on performance improvements and underlying representational changes. Using the standard EDT architecture \cite{wu2023elastic} as baseline, we compare against EDT-SIL and EDT-TIL variants across four continuous control tasks from the D4RL benchmark \cite{fu2020d4rl}: Ant, HalfCheetah, Hopper, and Walker2d. These locomotion tasks represent different movement challenges from quadrupedal (Ant) to bipedal locomotion (Hopper, Walker2d) and high-speed running (HalfCheetah), providing diverse sensorimotor dynamics for evaluating intrinsic motivation mechanisms.

We evaluated models on both medium datasets ($\sim$1M transitions with cleaner trajectories) and medium-replay datasets ($\sim$2M transitions with noisy replay buffer data), using five random seeds for statistical robustness. Performance was evaluated using Human-Normalized Scores (HNS), providing consistent scaling across environments: 
\begin{equation}
\label{HNS}
    \text{HNS} = \frac{\text{score} - \text{score}\_{\text{random}}}{\text{score}\_{\text{human}} - \text{score}\_{\text{random}}}
\end{equation}
For embedding analysis, we collected state embeddings during evaluation by executing the best performing model for each environment-dataset combination, extracting embeddings at each step over single episodes with maximum 1000 steps. From these embeddings, we computed three key geometric metrics (covariance trace, L2 norm, and cosine similarity) following our analysis framework, with results averaged across three repetitions for robustness.


\subsection{Performance Results}
\label{subsec: Performance Results}

Table \ref{tab: scores} presents performance results across both medium and medium-replay datasets, where intrinsic motivation variants demonstrate environment-specific effectiveness patterns. 
On medium datasets, EDT-TIL achieved the best performance in 2 out of 4 environments (Walker2d: 73.50 vs 68.50 HNS; Hopper: 59.63 vs 57.49/59.31 HNS for baseline/SIL).

The medium-replay datasets reveal different intrinsic motivation effectiveness patterns. EDT-SIL significantly outperforms the baseline in Hopper (84.67 vs 81.56 HNS), while EDT-TIL demonstrates robust performance in HalfCheetah (38.60 vs 37.32 HNS) and Walker2d (65.06 vs 62.25 HNS). Interestingly, the baseline EDT achieves the best performance in Ant on medium-replay (85.51 HNS), suggesting that this environment may be less prone to intrinsic motivation on noisier datasets. These results suggest that different intrinsic motivation mechanisms create complementary representational advantages suited to different environmental dynamics and dataset characteristics.

\begin{table}[h]
\caption{\footnotesize{Performance comparison on Medium and Medium-Replay datasets. Human-normalized scores (HNS) show mean ± standard deviation across 5 seeds. The best results per each environment are highlighted in bold.}}
\centering
\footnotesize
\begin{tabular}{@{}lcccc@{}}
\toprule
\textbf{Dataset/Model} & \textbf{Ant} & \textbf{HalfCheetah} & \textbf{Hopper} & \textbf{Walker2d} \\
\midrule
\textbf{Medium} \\
EDT & 88.84±3.61 & 42.30±0.14 & 57.49±3.81 & 68.50±2.03 \\
EDT-SIL & \textbf{90.49±5.01} & \textbf{42.46±0.12} & 59.31±6.16 & 69.44±4.46 \\
EDT-TIL & 89.01±5.83 & 42.18±0.34 & \textbf{59.63±2.35} & \textbf{73.50±4.29} \\
\midrule
\textbf{Medium-Replay} \\
EDT & \textbf{85.51±5.06} & 37.32±2.46 & 81.56±9.96 & 62.25±5.21 \\
EDT-SIL & 84.02±3.72 & 37.64±2.44 & \textbf{84.67±4.80} & 57.21±8.54 \\
EDT-TIL & 83.72±4.13 & \textbf{38.60±1.28} & 81.72±9.27 & \textbf{65.06±3.81} \\
\bottomrule
\end{tabular}
\label{tab: scores}
\end{table}




\subsection{Embedding Analysis Results}
\label{Embedding Analysis Results}

Table \ref{tab: environment embeddings comparison} reveals the mechanistic basis for performance improvements through analysis of embedding properties. Each environment exhibits a distinct correlation pattern between representational metrics and performance. \textbf{Ant} shows a strong negative correlation with covariance trace (-0.907), suggesting that reduced total variance distribution improves performance. \textbf{HalfCheetah} exhibits a positive correlation with covariance trace (0.850), indicating that increased representational capacity benefits this environment. \textbf{Hopper} demonstrates a positive correlation with cosine similarity (+0.658), suggesting that increased similarity between state representations enhances performance. \textbf{Walker2d} shows a strong negative correlation with cosine similarity (-0.950), indicating that increased orthogonality between embeddings is crucial. Such environment-specific patterns demonstrate that intrinsic motivation mechanisms create tailored representational structures aligned with task demands and consistent with the biological principle of adaptive representational organization.

Examining the embedding properties across models, we can further see and interesting effect. More specifically the analysis reveals distinct representational effects of each intrinsic motivation variant. 
EDT-SIL consistently creates more compact representations through reduced covariance trace and L2 norms. 
EDT-TIL promotes representational orthogonality via reduced cosine similarity (Walker2d: -0.950; Hopper: +0.658), demonstrating environment-specific optimization strategies that mirror biological neural decorrelation principles.
The complementary nature of those mechanisms suggests that different intrinsic motivation approaches implement distinct aspects of biological representational regulation. EDT-SIL enhances representational efficiency at the input level, while EDT-TIL optimizes sequential processing structures. This division of regulatory functions aligns with hierarchical organization principles observed in biological neural systems, where different processing stages maintain distinct homeostatic mechanisms. These findings may suggest that intrinsic motivation may help in acting as a representational prior, shaping the embedding geometry.


\begin{table*}[h]
\caption{\footnotesize{Performance and embedding properties comparison across environments. Best performance highlighted in bold. Strongest embedding-performance correlation indicated for each environment.}}
\centering
\small
\begin{tabular}{@{}lccccc|cccc@{}}
\toprule
\multirow{2}{*}{\textbf{Environment}} & \multirow{2}{*}{\textbf{Model}} & \multicolumn{5}{c}{\textbf{Performance \& Embeddings}} \\
\cmidrule{3-7}
& & \textbf{Performance} & \textbf{Cov. Trace} & \textbf{L2 Norm} & \textbf{Cos. Sim.} & \textbf{Correlation} \\
\midrule
\multirow{3}{*}{\textbf{Ant}} 
& Baseline & 88.84 & 620.76 & 24.92 & 0.0288 & \multirow{3}{*}{\parbox{2cm}{\centering Cov. Trace\\(r = -0.907)}} \\
& EDT-SIL & \textbf{90.49} & 526.00 & 23.19 & 0.0356 & \\
& EDT-TIL & 89.01 & 572.99 & 24.02 & 0.0278 & \\
\midrule
\multirow{3}{*}{\textbf{HalfCheetah}} 
& Baseline & 42.30 & 627.99 & 25.20 & 0.0270 & \multirow{3}{*}{\parbox{2cm}{\centering Cov. Trace\\(r = +0.850)}} \\
& EDT-SIL & \textbf{42.46} & 632.01 & 25.27 & 0.0272 & \\
& EDT-TIL & 42.18 & 563.11 & 24.12 & 0.0439 & \\
\midrule
\multirow{3}{*}{\textbf{Hopper}} 
& Baseline & 57.49 & 584.58 & 24.49 & 0.0795 & \multirow{3}{*}{\parbox{2cm}{\centering Cos. Sim.\\(r = +0.658)}} \\
& EDT-SIL & 59.31 & 508.99 & 23.08 & 0.0818 & \\
& EDT-TIL & \textbf{59.63} & 642.66 & 25.55 & 0.1167 & \\
\midrule
\multirow{3}{*}{\textbf{Walker2d}} 
& Baseline & 68.50 & 608.42 & 24.78 & 0.0811 & \multirow{3}{*}{\parbox{2cm}{\centering Cos. Sim.\\(r = -0.950)}} \\
& EDT-SIL & 69.44 & 523.33 & 23.33 & 0.0825 & \\
& EDT-TIL & \textbf{73.50} & 568.98 & 24.14 & 0.0731 & \\
\bottomrule
\end{tabular}
\label{tab: environment embeddings comparison}
\end{table*}



\section{Conclusions}
\label{sec:Conclusions}

In our work we investigated how introducing intrinsic motivation mechanisms in Elastic Decision Transformers shapes learned representations and their correlation with task performance across multiple continuous control environments. Our systematic analysis revealed several key findings that suggest the relationship between empirical performance improvements and underlying mechanisms.
Our experiments demonstrate that intrinsic motivation variants (EDT-SIL and EDT-TIL) consistently outperform the baseline EDT across most environments and datasets, with the 3-layer RND configuration emerging as optimal. The post-hoc explainability analysis reveals that each environment exhibits distinct patterns in how embedding properties correlate with performance. EDT-SIL creates compact representations through reduced covariance and L2 norms, while EDT-TIL promotes representational orthogonality through reduced cosine similarity, particularly evident in Walker2d.
These findings show that intrinsic motivation might operate as more than a simple exploration bonus: it acts as a representational prior that shapes embedding geometry in biologically plausible ways. The complementary nature of EDT-SIL and EDT-TIL mechanisms mirrors hierarchical organization principles observed in biological neural systems, where different processing stages might maintain distinct homeostatic mechanisms. The environment-specific correlation patterns suggest that intrinsic motivation mechanisms create tailored representational structures aligned with task demands, providing a mechanistic explanation for why intrinsic motivation improves performance beyond simple reward optimization.
Regarding the limitations of our study, we primarily focused on the geometrical properties of the embeddings rather than on explicit explainability measures. Further work could include application of explainability models to understand the impact of certain features and vector dimensionality reduction on the final performances of the models proposed. Moreover the initial variations in HNS show very small variations (of the order of percent). This is of course an initial assessment, which should be further exploited and tested, to further prove the significance of the results obtained.
In addition, future work could extend the framework to analyze transformer outputs and action representations, and investigate temporal dynamics of embedding evolution during training to understand how intrinsic motivation shapes learning trajectories over time.



\bibliographystyle{plain}
\bibliography{biblio}

@article{rnd,
  author       = {Yuri Burda and
                  Harrison Edwards and
                  Amos J. Storkey and
                  Oleg Klimov},
  title        = {Exploration by Random Network Distillation},
  journal      = {CoRR},
  volume       = {abs/1810.12894},
  year         = {2018},
  url          = {http://arxiv.org/abs/1810.12894},
  eprinttype    = {arXiv},
  eprint       = {1810.12894},
  bibsource    = {dblp computer science bibliography, https://dblp.org}
}

@inproceedings{esann2025,
  title={Introducing intrinsic motivation in elastic decision transformers},
  author={Guiducci, Leonardo and Dimitri, Giovanna Maria and Palma, Giulia and Rizzo, Antonio},
  booktitle={ESANN 2025 proceedings, European Symposium on Artificial Neural Networks, Computational Intelligence and Machine Learning},
  year={2025},
  address={Bruges (Belgium) and online event},
  month={23--25~} # apr,
  publisher={i6doc.com publ.},
  isbn={9782875870933},
  url={http://www.i6doc.com/en/}
}

@inproceedings{chen2021decision,
  title={Decision Transformer: Reinforcement Learning via Sequence Modeling},
  author={Chen, Lili and Lu, Kevin and Rajeswaran, Aravind and Lee, Kimin and Grover, Aditya and Laskin, Michael and Abbeel, Pieter and Srinivas, Aravind and Mordatch, Igor},
  booktitle={Advances in Neural Information Processing Systems},
  volume={34},
  pages={15084--15097},
  year={2021}
}

@article{oudeyer2007intrinsic,
  title={What is intrinsic motivation? A typology of computational approaches},
  author={Oudeyer, Pierre-Yves and Kaplan, Frederic},
  journal={Frontiers in neurorobotics},
  volume={1},
  pages={108},
  year={2007},
  publisher={Frontiers}
}

@inproceedings{transformer,
 author = {Vaswani, Ashish and Shazeer, Noam and Parmar, Niki and Uszkoreit, Jakob and Jones, Llion and Gomez, Aidan N and Kaiser, \L ukasz and Polosukhin, Illia},
 booktitle = {Advances in Neural Information Processing Systems},
 editor = {I. Guyon and U. Von Luxburg and S. Bengio and H. Wallach and R. Fergus and S. Vishwanathan and R. Garnett},
 pages = {},
 publisher = {Curran Associates, Inc.},
 title = {Attention is All you Need},
 url = {https://proceedings.neurips.cc/paper_files/paper/2017/file/3f5ee243547dee91fbd053c1c4a845aa-Paper.pdf},
 volume = {30},
 year = {2017}
}

@inproceedings{wu2023elastic,
  title={Elastic Decision Transformer},
  author={Wu, Yueh-Hua and Wang, Xiaolong and Hamaya, Masashi},
  booktitle={Advances in Neural Information Processing Systems},
  volume={36},
  year={2023},
  url={https://arxiv.org/abs/2307.02484}
}

@InProceedings{zenke2017,
  title={Continual Learning Through Synaptic Intelligence},
  author={Friedemann Zenke and Ben Poole and Surya Ganguli},
  booktitle={Proceedings of the 34th International Conference on Machine Learning},
  pages={3987--3995},
  year={2017},
  editor={Precup, Doina and Teh, Yee Whye},
  volume={70},
  series={Proceedings of Machine Learning Research},
  month={06--11 Aug},
  publisher={PMLR},
  url={https://proceedings.mlr.press/v70/zenke17a.html}
}

@article{Clark_2013, 
title={Whatever next? Predictive brains, situated agents, and the future of cognitive science}, 
volume={36}, 
DOI={10.1017/S0140525X12000477}, 
number={3}, 
journal={Behavioral and Brain Sciences}, 
author={Clark, Andy}, 
year={2013}, 
pages={181–204}}

@article{ganguli2012compressed,
  title={Compressed sensing, sparsity, and dimensionality in neuronal information processing and data analysis},
  author={Ganguli, Surya and Sompolinsky, Haim},
  journal={Annual review of neuroscience},
  volume={35},
  number={1},
  pages={485--508},
  year={2012},
  publisher={Annual Reviews}
}

@article{friston2010free,
  title={The free-energy principle: a unified brain theory?},
  author={Friston, Karl},
  journal={Nature Reviews Neuroscience},
  volume={11},
  number={2},
  pages={127--138},
  year={2010},
  publisher={Nature Publishing Group},
  doi={10.1038/nrn2787}
}

@article{friston2009predictive,
  title={Predictive coding under the free-energy principle},
  author={Friston, Karl and Kiebel, Stefan},
  journal={Philosophical Transactions of the Royal Society B: Biological Sciences},
  volume={364},
  number={1521},
  pages={1211--1221},
  year={2009},
  month={may},
  publisher={The Royal Society},
  doi={10.1098/rstb.2008.0300},
  pmid={19528002},
  pmcid={PMC2666703}
}

@article{allostasi1988,
author = {Sterling, Peter and Eyer, Joseph},
year = {1988},
month = {01},
pages = {},
title = {Allostasis: A New Paradigm to Explain Arousal Pathology},
journal = {Handbook of Life Stress, Cognition and Health}
}

@inproceedings{pathak2017curiosity,
  title={Curiosity-driven Exploration by Self-supervised Prediction},
  author={Pathak, Deepak and Agrawal, Pulkit and Efros, Alexei A and Darrell, Trevor},
  booktitle={Proceedings of the IEEE Conference on Computer Vision and Pattern Recognition Workshops},
  year={2017}
}

@article{fu2020d4rl,
  title={D4rl: Datasets for deep data-driven reinforcement learning},
  author={Fu, Justin and Kumar, Aviral and Nachum, Ofir and Tucker, George and Levine, Sergey},
  journal={arXiv preprint arXiv:2004.07219},
  year={2020}
}

@article{levine2020offline,
  title={Offline reinforcement learning: Tutorial, review, and perspectives on open problems},
  author={Levine, Sergey and Kumar, Aviral and Tucker, George and Fu, Justin},
  journal={arXiv preprint arXiv:2005.01643},
  year={2020}
}

@inproceedings{kumar2020conservative,
  title={Conservative Q-learning for offline reinforcement learning},
  author={Kumar, Aviral and Zhou, Aurick and Tucker, George and Levine, Sergey},
  booktitle={Advances in Neural Information Processing Systems},
  volume={33},
  year={2020}
}

@inproceedings{kidambi2020morel,
  title={Morel: Model-based offline reinforcement learning},
  author={Kidambi, Raghavendra P and Rajeswaran, Aravind and Netrapalli, Praneeth and Joachims, Thorsten},
  booktitle={Advances in Neural Information Processing Systems},
  volume={33},
  year={2020}
}

@article{bengio2013representation,
  title={Representation learning: A review and new perspectives},
  author={Bengio, Yoshua and Courville, Aaron and Vincent, Pascal},
  journal={IEEE transactions on pattern analysis and machine intelligence},
  volume={35},
  number={8},
  pages={1798--1828},
  year={2013},
  publisher={IEEE}
}


\newpage
\appendix



\section{Biological Plausibility}
\label{app:bio_plaus}

In this section we propose some insights on the biological plausibility of the proposed model.

\subsection{Homeostatic Regulation: towards a biologically inspired reinforcement learning approach}
\label{subsec: biological plausibility and homeostatic regulation}

 The quest for explainable AI in reinforcement learning (RL) has a tight parallel with biological learning, where intrinsic motivation shapes adaptive behavior through allostatic regulation, achieving stability by adjusting predictions rather than fixing a--priori parameters \cite{allostasi1988}. Such predictive adaptation is reflected in how the so called Random Network Distillation (RND) models operate: learning is driven by discrepancies between predicted and actual inputs, echoing brain mechanisms that constantly update internal models based on sensory prediction errors \cite{friston2010free, Clark_2013}. These prediction hierarchies span from primary sensory to higher-order cortical processing \cite{friston2009predictive}, suggesting intrinsic motivation can be applied across representational levels. Moreover, biological systems maintain representational homeostasis, optimizing information processing through the regulation of capacity and structural organization \cite{zenke2017, ganguli2012compressed}. Intrinsic motivation fosters flexible learning and generalization \cite{oudeyer2007intrinsic}. In transformer-based models like Elastic Decision Transformers, auxiliary losses based on RND act as allostatic regulators, guiding representational structure without altering offline reward signals. Such losses can help preventing representational collapse, mirroring biological mechanisms that sustain learning adaptability and predictive efficiency.



\section{Random Network Distillation}
\label{app:RND}

In this section, we detail how the RND module is integrated into the EDT architecture. We further present the analysis conducted to find the optimal number of layers in RND networks.


\subsection{RND Architecture Analysis}
\label{app: rnd network architecture analysis}

\begin{figure*}[h]
\centering
\includegraphics[width=0.8\textwidth]{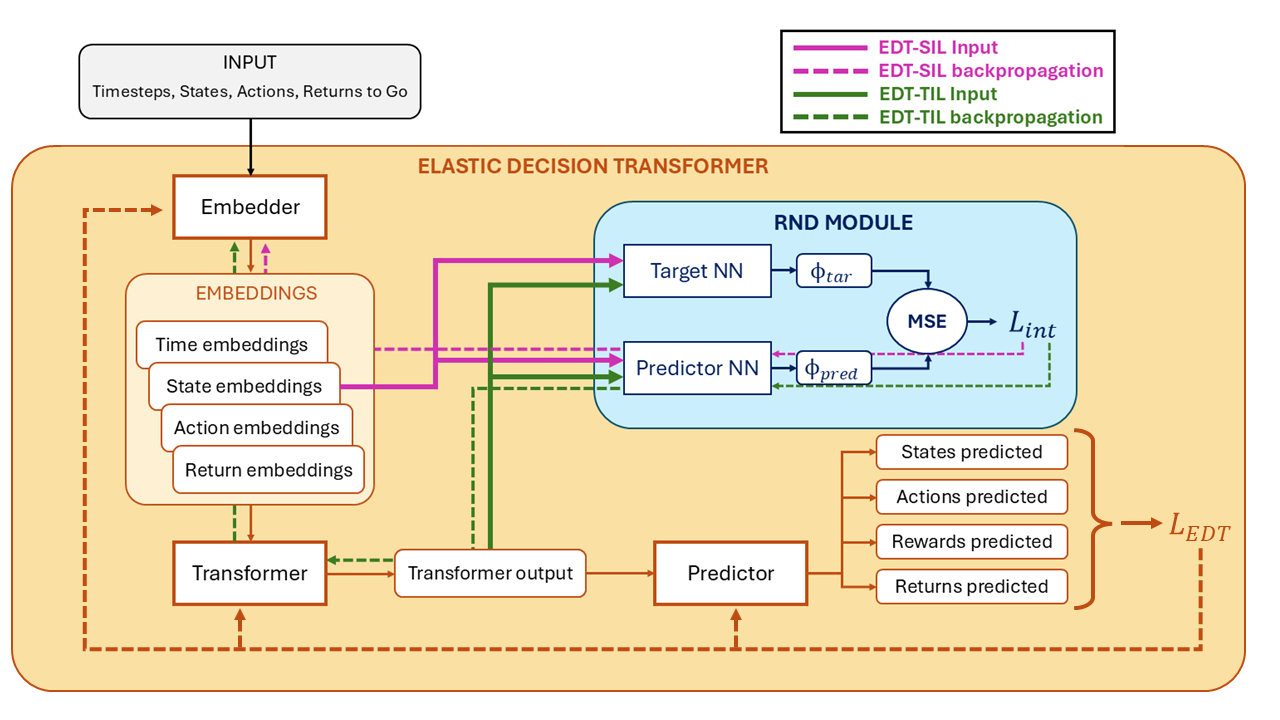}
\caption{\footnotesize{Architecture of the Elastic Decision Transformer with intrinsic motivation mechanisms proposed in \cite{esann2025}.}}
\label{fig: edt architecture}
\end{figure*}

Figure \ref{fig: edt architecture} shows both EDT-SIL and EDT-TIL variants, where the RND module operates on state embeddings or transformer outputs respectively. The dashed lines indicate backpropagation paths for each variant, with $L_{int}$ from the RND module contributing to the total loss $L_{EDT}$ alongside standard prediction losses. The target network of the RND module has frozen weights and is never updated, as proposed in \cite{rnd}. 

\subsection{RND Layer Configuration Analysis}
\label{app:rnd_analysis}

To optimize the intrinsic motivation mechanism and understand the impact of RND network capacity on intrinsic motivation effectiveness, we conducted a systematic investigation of the predictor network depth on both performance and representation quality. The motivation for this analysis stems from the hypothesis that different network capacities may capture different levels of representational complexity, potentially affecting both the quality of intrinsic rewards and the resulting policy performance. We evaluated three RND predictor configurations: a 1-layer RND (considering it as a minimal architecture to establish a baseline for intrinsic reward generation), a 3-layers RND as the default configuration in \cite{esann2025}, and a 10-layers RND, considering it as a high-capacity variant to test whether increased expressiveness improves intrinsic motivation. This analysis was conducted exclusively on our best performing dataset (\textit{i.e.} Medium Datasets), as these demonstrated the most promising initial results, in order to optimize the RND predictor network architecture.

\begin{figure}[h]
\centering
\includegraphics[width=\textwidth]{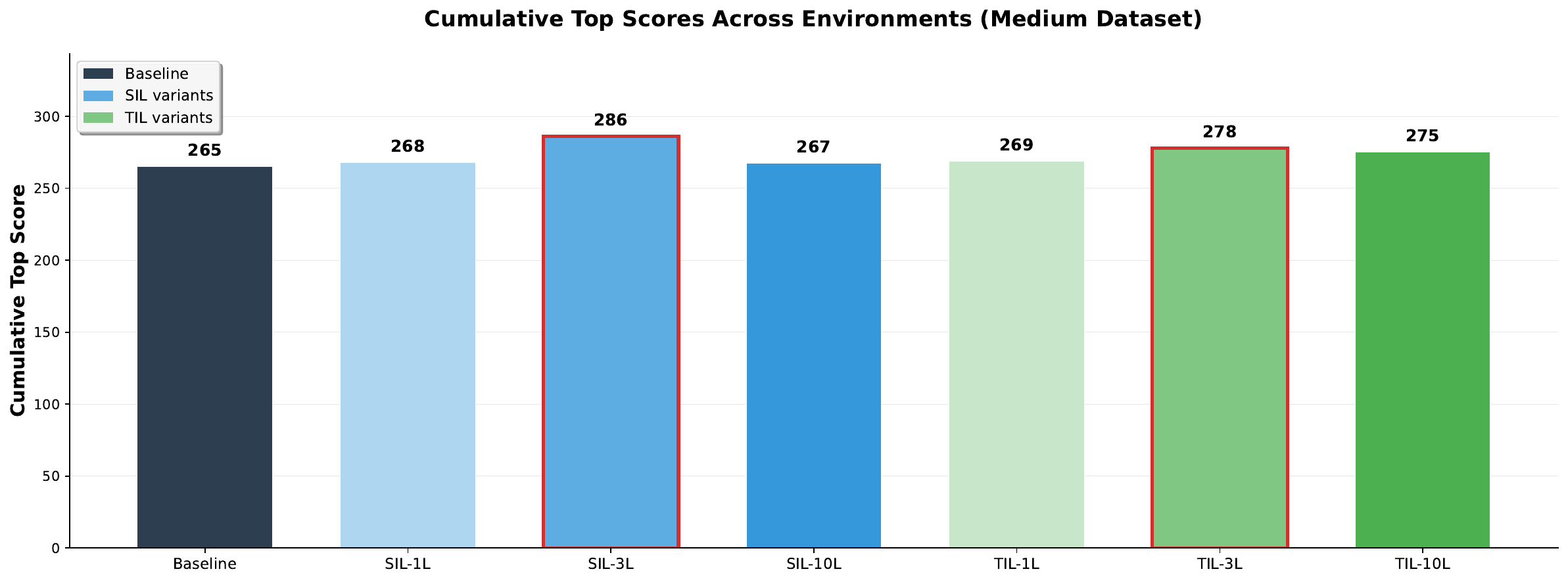}
\caption{\footnotesize{Cumulative Human-Normalized Scores (HNS) obtained by the best models trained on each environment of the Medium dataset. 3-layer variants (red borders) achieve optimal performance for both SIL and TIL mechanisms.}}
\label{fig:cumulative_top_scores}
\end{figure}

The 3-layer configuration emerged as optimal across both EDT-SIL and EDT-TIL variants. Figure \ref{fig:cumulative_top_scores} demonstrates that 3-layer variants (highlighted with red borders) consistently achieve the highest cumulative scores across all environments. This finding aligns with the biological principle of representational balance: too few layers (1-layer) may lack sufficient capacity to capture complex predictive relationships, while too many layers (10-layer) may lead to overfitting or representational instability.

\section{Extended Embedding Analysis}
\label{app:embedding_analysis}

In this section, we detail the metrics used to analyze embeddings and provide an in-depth analysis of the properties of embeddings in medium-replay datasets.


\subsection{Embedding Characterization Framework}
\label{app: embedding characterization framework}

We focused our analysis on three key geometric properties that capture different aspects of representational structure:

\begin{itemize}
    \item \textbf{Covariance Trace:} This metric measures the total variance distributed across embedding dimensions as per:
    \begin{equation}
    \label{eq: cov_trace}
        cov\_trace = \mathrm{Tr}(\mathrm{Cov}(E))
    \end{equation}
    where $E \in \mathbb{R}^{N \times d}$ represents the embedding matrix. By definition, the trace equals the sum of variances along each dimension, indicating how much total information is captured across the representational space.

    \item \textbf{L2 Norm:} The mean magnitude of embedding vectors quantifies representational compactness, as per:
    \begin{equation}
    \label{eq: l2_norm}
    l2\_norm = \frac{1}{N} \sum_{i=1}^{N} |e_i|_2
    \end{equation}
    This metric provides insight into the average energy or magnitude of embedding vectors in the representational space.
    
    \item \textbf{Cosine Similarity:} The average pairwise cosine similarity within each tensor assesses representational orthogonality as per:
    \begin{equation}
    \label{eq: cosine_sim}
        \mathrm{cos\_sim} = \frac{1}{|\mathcal{P}|} \sum_{(e_i, e_j) \in \mathcal{P}} \frac{e_i \cdot e_j}{|e_i||e_j|}
    \end{equation}
    where $\mathcal{P}$ are made of all pairs of embedding vectors within each tensor. Lower values suggest more orthogonal representations, which may indicate better disentanglement of different aspects of the state space.
\end{itemize}

\subsection{Detailed Medium-Replay Dataset Analysis}
\label{app:medium_replay}

The main paper focused on medium dataset results. Table \ref{tab:medium_replay_detailed} provides comprehensive embedding analysis for medium-replay datasets, showing how noise affects representational properties.

\begin{table}[h]
\centering
\caption{Embedding properties on medium-replay dataset}
\label{tab:medium_replay_detailed}
\small
\begin{tabular}{llrrrr}
\toprule
Environment & Model & Performance & cov\_trace & l2\_norm & cosine\_sim \\
\midrule
\multirow{3}{*}{Ant} & Baseline & \textbf{85.51} & 582.17 & 24.27 & 0.0262 \\
 & SIL-3L & 84.02 & 532.72 & 23.46 & 0.0427 \\
 & TIL-3L & 83.72 & 578.72 & 24.20 & 0.0273 \\
\addlinespace
\multirow{3}{*}{HalfCheetah} & Baseline & 37.32 & 622.22 & 25.20 & 0.0311 \\
 & SIL-3L & 37.64 & 564.79 & 24.20 & 0.0427 \\
 & TIL-3L & \textbf{38.60} & 620.17 & 25.15 & 0.0296 \\
\addlinespace
\multirow{3}{*}{Hopper} & Baseline & 81.56 & 595.53 & 24.90 & 0.0888 \\
 & SIL-3L & \textbf{84.67} & 530.42 & 23.80 & 0.1178 \\
 & TIL-3L & 81.72 & 605.05 & 25.10 & 0.0973 \\
\addlinespace
\multirow{3}{*}{Walker2d} & Baseline & 62.25 & 589.11 & 24.72 & 0.0905 \\
 & SIL-3L & 57.21 & 524.62 & 23.53 & 0.0957 \\
 & TIL-3L & \textbf{65.06} & 571.08 & 24.31 & 0.0786 \\
\bottomrule
\end{tabular}
\end{table}

Comparing Table \ref{tab:medium_replay_detailed} with Table \ref{tab: environment embeddings comparison} in the main paper reveals noise-specific effects:
\begin{enumerate}
    \item \textbf{Ant}: Baseline recovers superiority, suggesting SIL's compactness is sensitive to noise
    \item \textbf{HalfCheetah}: TIL maintains advantage, showing robustness to noise
    \item \textbf{Walker2d}: TIL's orthogonality benefits persist even with noise
\end{enumerate}



\section{Statistical Analysis Details}
\label{app:statistical_details}


\subsection{ANOVA Results}
\label{app:anova}

Beyond the correlation analysis in the main paper, we conducted ANOVA tests to assess statistical significance of embedding differences between models.

\begin{table}[h]
\centering
\caption{ANOVA results for embedding metrics (selected significant results)}
\label{tab:anova_results}
\begin{tabular}{llrr}
\toprule
Environment & Metric & F-statistic & $p$-value \\
\midrule
\multicolumn{4}{c}{\textbf{Medium Dataset}} \\
\midrule
Ant & cov\_trace & 10.69 & 0.011 \\
 & l2\_norm\_mean & 11.25 & 0.009 \\
HalfCheetah & cov\_trace & 380.16 & $<$0.001 \\
 & cosine\_similarity\_mean & 701.46 & $<$0.001 \\
Hopper & cov\_trace & 351.71 & $<$0.001 \\
 & l2\_norm\_mean & 325.74 & $<$0.001 \\
Walker2d & cosine\_similarity\_mean & 56.16 & $<$0.001 \\
\midrule
\multicolumn{4}{c}{\textbf{Medium-Replay Dataset}} \\
\midrule
HalfCheetah & cov\_trace & 973.83 & $<$0.001 \\
Hopper & l2\_norm\_mean & 53.87 & $<$0.001 \\
Walker2d & cosine\_similarity\_mean & 15.34 & 0.004 \\
\bottomrule
\end{tabular}
\end{table}

Table \ref{tab:anova_results} confirms that the embedding differences observed are statistically significant, with F-statistics indicating large effect sizes for most comparisons.



\section{Implementation Details}
\label{app:implementation}


\subsection{Training Configuration}
\label{app:training_config}

\textbf{Complete Loss Function:}
\begin{equation}
    \mathcal{L}_{total} = \mathcal{L}_{action} + \alpha\mathcal{L}_{state} + \beta\mathcal{L}_{exp} + \gamma\mathcal{L}_{ret} + \mathcal{L}_{int}
\end{equation}

where $\mathcal{L}_{action}$ is the MSE loss for action prediction, $\mathcal{L}_{state}$ is the MSE loss for next state prediction, $\mathcal{L}_{exp}$ is the expectile regression loss for return estimation, $\mathcal{L}_{ret}$ is the cross-entropy loss for discretized returns, and $\mathcal{L}_{int}$ is the intrinsic loss from RND. The weights are $\alpha = 0.1$, $\beta = 1.0$, and $\gamma = 0.001$.

\textbf{Hyperparameters:}
\begin{itemize}
    \item Optimizer: AdamW with $lr = 10^{-4}$, weight decay $10^{-4}$
    \item Batch size: 256 
    \item Gradient clipping: 0.25
    \item Expectile level: $\alpha = 0.99$
    \item Context length: 20
\end{itemize}

The 3-layer RND configuration consists of a predictor network (Linear(input\_size, 512) $\rightarrow$ ELU $\rightarrow$ Linear(512, 512) $\rightarrow$ ELU $\rightarrow$ Linear(512, 512)) and a target network (Linear(input\_size, 512)). All linear layers use orthogonal initialization with gain $\sqrt{2}$ and zero bias initialization as in \cite{rnd}. The target network remains frozen during training. Training requires approximately 1 hour and 3GB VRAM on an NVIDIA RTX 2080Ti system with Intel i9-9820X processor and 64GB RAM. EDT-SIL and EDT-TIL variants introduce negligible computational overhead compared to baseline EDT.

Following standard D4RL protocol, we use the complete offline datasets for training without traditional train/validation splits, as these datasets are specifically curated for offline RL evaluation. Model performance is assessed by deploying trained policies in the MuJoCo environments for 100 episodes, rather than on held-out trajectory data.


\end{document}